\documentclass[letterpaper, 10 pt, conference]{ieeeconf}
\IEEEoverridecommandlockouts
\usepackage{amsmath,amsfonts}
\usepackage{algorithmic}
\usepackage{algorithm}
\usepackage{array}
\usepackage[caption=false,font=normalsize,labelfont=sf,textfont=sf]{subfig}
\usepackage{textcomp}
\usepackage{stfloats}
\usepackage{url}
\usepackage{verbatim}
\usepackage{graphicx}
\usepackage{cite}
\usepackage{tikz}
\usepackage{graphicx}
\usepackage{adjustbox}
\usepackage{capt-of}
\usepackage{booktabs}

\begin{document}

\title{\textbf{MoMo}: Dial \textbf{Mo}tion \textbf{Mo}de in Robot Manipulation with Spatiotemporal Action Tokenization}

\author{Yuhan Hu, Hugues Thomas, Peide Huang, Mouli Sivapurapu, Benoit Landry, Arto Kivila ~\IEEEmembership{Apple}}

\IEEEaftertitletext{%
  \noindent\begin{minipage}{\textwidth}
    \centering
    \includegraphics[width=\textwidth]{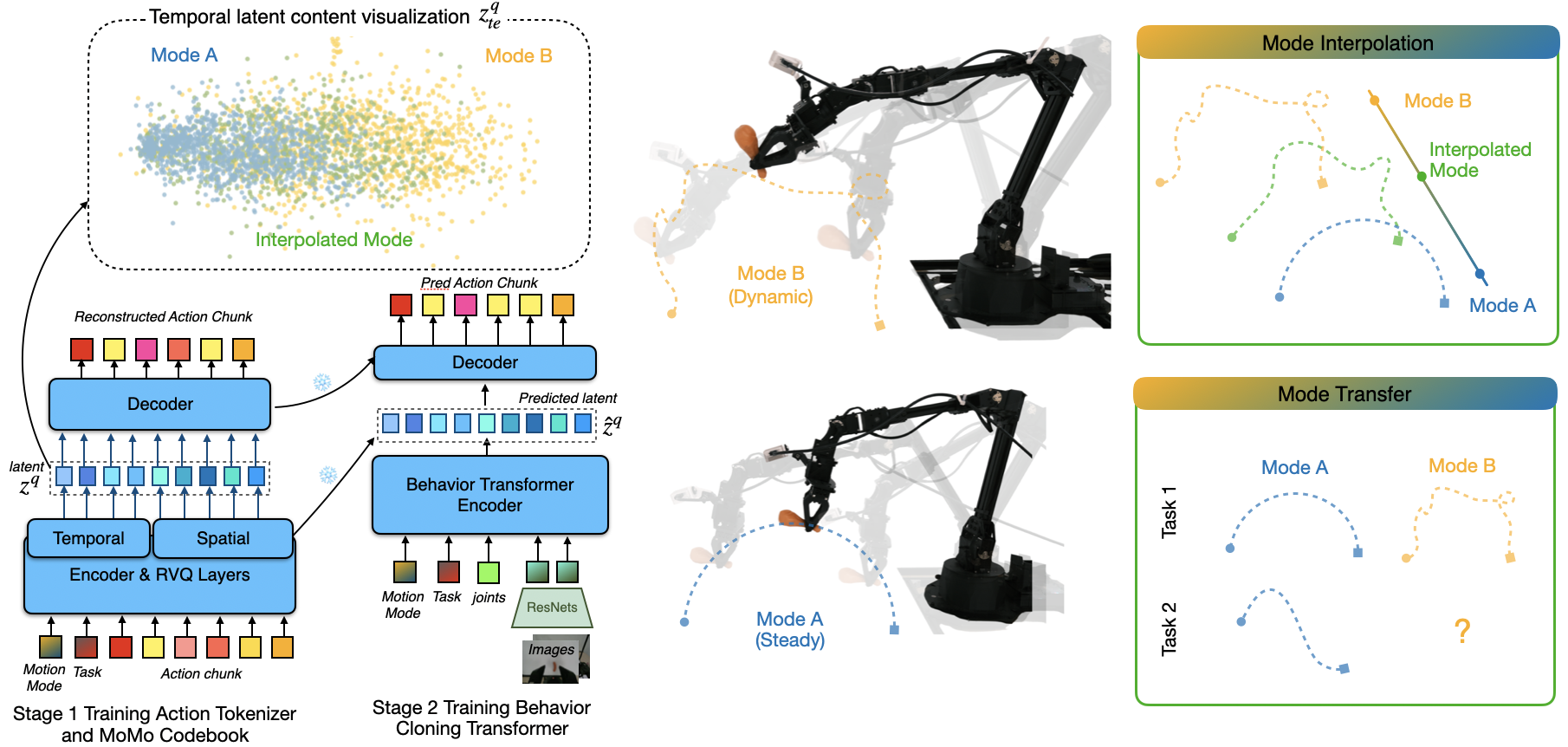}
    \captionof{figure}{Overview of MoMo (architecture detailed in Figure~\ref{fig:arch}). \textbf{Left:} the Stage~1/Stage~2 pipeline, with a temporal latent visualization (top) showing Mode A, Mode B, and intermediate-mode codes forming an ordered band. \textbf{Center:} example rollouts of the same task in Mode B (dynamic) and Mode A (steady). \textbf{Right:} the intermediate mode query at inference (top) and compositional mode transfer to an unseen (task, mode) pair (bottom).}
    \label{fig:teaser}
  \end{minipage}%
  \vspace{\baselineskip}%
}

\maketitle
\thispagestyle{empty}
\pagestyle{empty}

\begin{abstract}
To operate effectively across diverse contexts, robots must not only perform manipulation tasks accurately but also adapt how their actions unfold to the task, object, and interaction setting. We ask whether this execution-level variation can be learned as a reusable behavioral factor shared across tasks. We present \textbf{MoMo}, a two-stage imitation-learning framework consisting of a spatiotemporal action tokenizer and a behavior-cloning transformer that takes task and a continuous motion-mode condition as inputs. Across six real-robot manipulation tasks, varying this condition produces steady, dynamic, and intermediate behaviors that human raters can distinguish and that differ in joint speed, acceleration, and end-effector approach pitch. On tasks demonstrated in only one mode, MoMo transfers the unseen requested mode while largely preserving task success. Together, these results provide evidence of compositional generalization to unseen task--mode combinations and show that motion mode can be reused across tasks to control how a manipulation skill is performed.
\end{abstract}


\section{Introduction}

For robots to operate effectively around humans---taking instructions, responding, or executing commands---they must not only perform actions accurately but also adapt \emph{how} those actions are executed to suit the task, object, and interaction context. This raises a broader question for imitation learning: can manipulation policies be conditioned on reusable behavioral factors, rather than relearning that behavior separately for every task?

We study this question through \emph{motion mode}: a task-agnostic behavioral modulation factor that changes how a manipulation task unfolds while leaving its functional goal unchanged. Motion mode is not a skill primitive---it does not specify a task such as reaching, grasping, or pouring by itself. Instead, it modulates an underlying task policy along execution-quality dimensions such as pacing, smoothness, hesitation, acceleration, and end-effector orientation.

Existing behavior- and style-conditioned manipulation policies typically learn the behavior jointly with each task, so extending a behavior to a new task requires demonstrations of that task in that behavior; the behavior never becomes a factor that can be carried across tasks. Our approach is motivated by action tokenization in imitation learning~\cite{shafiullah2022behavior,lee2024behavior,pertsch2024fast} and by prior work on disentangled motion representations in animated characters~\cite{zargarbashi2026vqstyle}. Together, these suggest that discrete action codes may provide a useful interface for separating task-relevant action structure from reusable behavioral variation. We bring this idea to robot manipulation and ask whether a motion-mode factor can be represented, controlled, and reused across tasks in a learned action-token space.

We propose \textbf{MoMo} (Figure~\ref{fig:teaser}), a two-stage imitation-learning framework that decomposes each action chunk into complementary spatial and temporal code streams. The spatial stream operates directly on the joint-space trajectory and emphasizes task-dependent structure such as path geometry, contact, and end-effector goals. The temporal stream operates on DCT features of joint deltas and emphasizes motion-mode qualities such as pacing, acceleration, smoothness, and hesitation. We instantiate two prototypes, steady Mode~A and dynamic Mode~B (\S\ref{sec:motionmode}). Stage~1 quantizes the spatial and temporal streams with separate RVQ stacks, then fuses them for reconstruction so task and behavioral variation are organized separately but decoded jointly. This decomposition provides an explicit interface for combining task-dependent action structure with behavioral variation that can be reused across tasks.

In Stage~2, a behavior-cloning transformer predicts the Stage~1 code indices from visual observations, proprioception, a task identifier, and a continuous motion-mode scaler; the frozen tokenizer then fuses and decodes the predicted spatial and temporal codes into executable action chunks. We evaluate six real-robot manipulation tasks: four are demonstrated in both modes and two in a single mode, creating held-out, out-of-distribution (OOD) task--mode pairs.

At inference the policy is conditioned jointly on the task identifier and the motion-mode scaler, letting us query task--mode combinations including ones never demonstrated. Sweeping the scaler produces behaviors that interpolate between the two prototypes, are distinguishable to blinded human raters, and differ in joint speed, joint acceleration, and end-effector approach pitch. In latent space, spatial codes group mainly by task and temporal codes by mode; on held-out pairs the model keeps the spatial codes inside the task cluster while shifting the temporal codes toward the requested mode's region of the training distribution, and execution moves toward the requested mode with task completion largely preserved. Together, these results provide evidence of compositional generalization to unseen task--mode pairs and show that motion mode can act as a reusable, task-agnostic behavioral control at inference time (videos in the supplementary material).

\section{Related Work}

\subsection{Imitation Learning for Robot Manipulation}
Imitation learning is a central paradigm for teaching robots manipulation from human demonstrations~\cite{argall2009survey}; recent work favors transformer architectures over long-horizon action chunks, from ACT~\cite{zhao2023learning} to diffusion policies~\cite{chi2023diffusion}, hierarchical extensions~\cite{park2024hierarchical}, Perceiver-Actor~\cite{shridhar2023perceiver}, and large-scale data scaling~\cite{lin2024data}. We build on this foundation but represent actions discretely, following the action-tokenization literature below.

\subsection{Action Tokenization for Imitation Policies}
A growing body of work treats robot actions as a discrete vocabulary predicted by a transformer: BeT~\cite{shafiullah2022behavior} clusters actions with k-means, VQ-BeT~\cite{lee2024behavior} replaces this with a residual VQ-VAE that decouples the tokenizer from the policy, and FAST~\cite{pertsch2024fast} tokenizes actions in the DCT domain via scalar quantization and byte-pair encoding. Our work draws on both threads---a learned RVQ tokenizer as in VQ-BeT, DCT features as in FAST, computed over joint \emph{deltas}---but unlike either, we maintain two parallel codebook stacks, one spatial and one temporal, enabling compositional control.

\subsection{Motion-Mode Disentanglement and Transfer}
While most imitation-learning frameworks target task success, a parallel literature encodes and transfers motion characteristics, often called ``style.'' In robotics, this includes Neural Policy Style Transfer~\cite{fernandez2023transferring} and constrained mode learning from imperfect demonstrations~\cite{wen2025constrained}. In animation, style transfer has been studied via adversarial disentanglement~\cite{fares2023zero}, meta-networks~\cite{pan2021fast}, diffusion~\cite{song2024arbitrary}, body-part decomposition~\cite{jang2022motion}, and video-to-animation transfer~\cite{aberman2020unpaired}; most relevant, Zargarbashi et al.~\cite{zargarbashi2026vqstyle} train residual VQ codebooks for character locomotion with contrastive losses so deeper codebook levels align with motion characteristics. We adapt this line to manipulation, where the motion-mode manifold is lower-dimensional and intertwined with environment-driven path variation, via a temporal/spatial split, residual contrastive loss, and matched same-task opposite-mode swap supervision.

\subsection{Robot Motion Characteristics and Legibility}
Beyond functional success, research on robot motion characteristics has emphasized its role in communication, task adaptation, and safety: legibility and predictability of intent~\cite{dragan2013legibility}, a design framework separating task goals from trajectory characteristics~\cite{hu2025elegnt}, and high-level control over motion via language prompts~\cite{mahadevan2024generative} or timing~\cite{zhou2017expressive}. These studies motivate an architecture in which task content and motion mode are separately addressable.

\section{Characterizing Motion Mode}
\label{sec:motionmode}

\begin{figure}
\centering
\includegraphics[width=\linewidth]{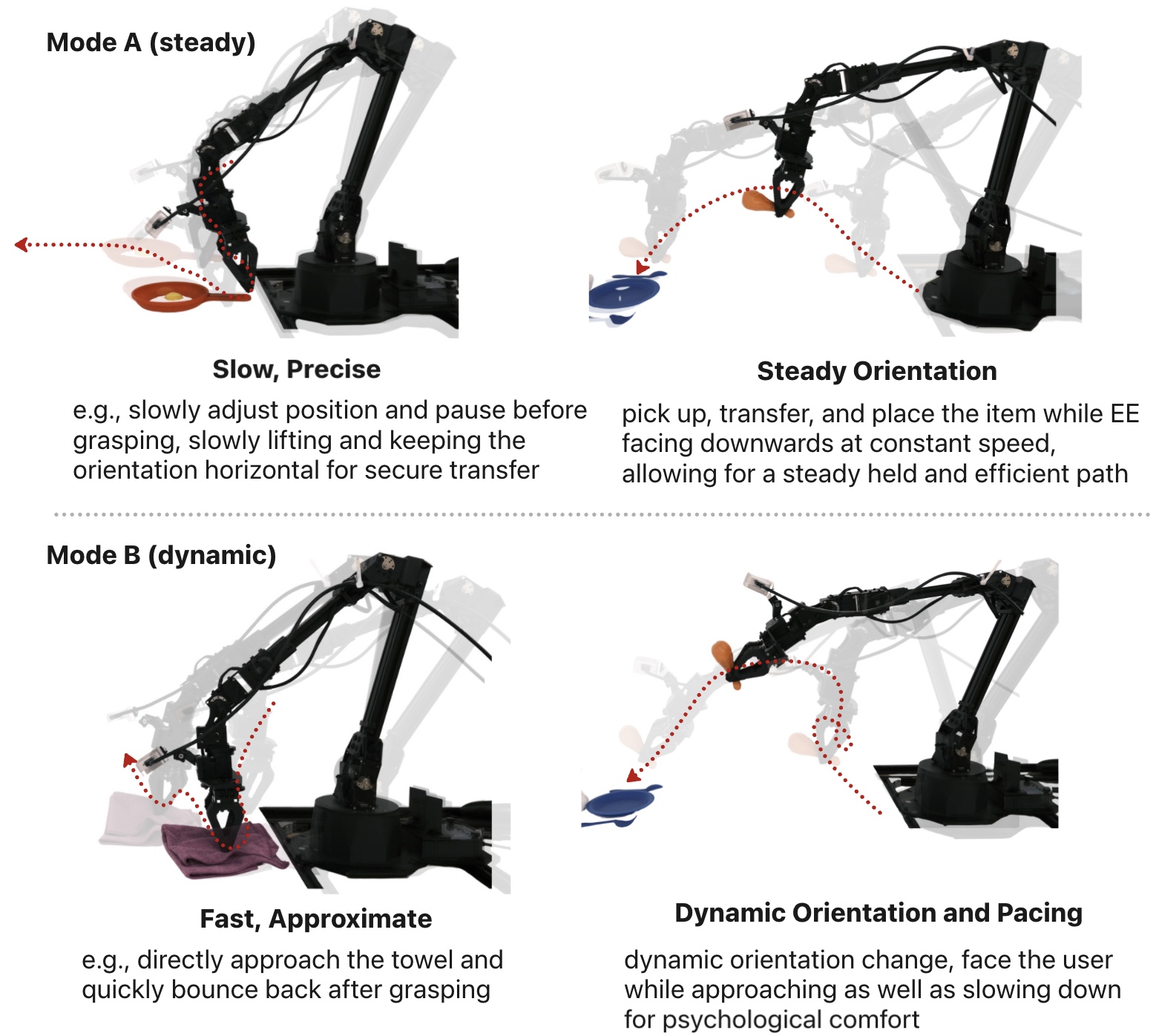}
\caption{Mode A vs.\ Mode B on real rollouts, illustrating why each suits different manipulation contexts. \textbf{Mode A} (steady): slow, precise motion that holds a constant end-effector orientation, suited to secure transfer of delicate items. \textbf{Mode B} (dynamic): faster, more tolerant pacing with dynamic orientation changes (e.g., facing the user), suited to interaction-driven handling.}
\label{fig:trajectory}
\end{figure}

We refer to trajectory characteristics beyond the functional objective as \textbf{motion modes}---the layer of variation in \emph{how} a task is executed rather than \emph{what} is executed (Figure~\ref{fig:trajectory}). Accordingly, a motion mode is not a task label or an independent skill, but a candidate behavioral factor whose meaning should remain consistent when applied to different tasks. These variations are of two kinds. \emph{Spatial} properties---path curvature, end-effector orientation---are set by the geometric path and depend weakly on how fast it is traversed; \emph{temporal} properties---pacing, smoothness, and pausing---are set by how that path is parameterized over time and depend weakly on its geometry. Motion modes shape how users interpret a robot's capability, reliability, and efficiency, and improve legibility~\cite{dragan2013legibility,hu2025elegnt}. We focus primarily on the temporal axis, along which our prototypes differ most strongly and which we manipulate throughout, though a smaller spatial component persists alongside it, as we quantify below.

For our study we adopt two opposing prototypes: \emph{Mode A} (direct, steady, slow), with low velocity variance and uniform acceleration---adjusting position slowly, pausing before grasping, and keeping the gripper level along an efficient path; and \emph{Mode B} (dynamic, varied), with higher speed peaks, more acceleration, and small bounce-back or hesitation motions, including interaction-driven changes such as turning to face a user mid-approach and slowing only as it nears them. We assign a \emph{motion-mode scaler} $s$ to each demonstration---low for Mode~B and high for Mode~A---and query the policy at one intermediate value between them. This two-prototype design is chosen for tractability; real motion-mode variation is richer and would likely need a higher-dimensional conditioning signal. The two modes differ primarily in pacing but also carry a spatial correlate: the gripper's approach angle differs systematically between them, which we later quantify with an approach-pitch metric from forward kinematics (\S\ref{sec:exp_real})---evidence that motion mode is predominantly, but not exclusively, temporal. This split motivates the dual-branch architecture: one branch for the path's geometry, the other for its temporal structure.

\section{Methodology}

\subsection{System Setup}

We use off-the-shelf Interbotix X-Series hardware in a single-arm ALOHA leader-follower configuration~\cite{zhao2023learning}: a WidowX-250 leader arm and a ViperX-300 6-DOF follower, used for both teleoperated data collection and policy execution. Two cameras capture the scene---a downward-facing fixed camera above the platform and a wrist camera on the gripper---both at $480 \times 640$. Data collection and execution run at 50~Hz, and action chunks of length $T = 100$ span 2~seconds of motion. The testbed is a tabletop with a preparation zone and a target zone of plates and utensils; tasks move objects from the former to the latter under different motion modes.

\subsection{Data Collection}

We consider tabletop manipulation with objects such as fake food, plates, pans, pots, and tissues. For each task, an operator demonstrates both the Mode~A and Mode~B prototypes to achieve the same functional goal, and each trajectory is recorded as video and annotated with the motion-mode scaler $s$. Initial object positions in the preparation zone and goal positions in the target zone are fixed (both within $\leq 5$~cm), and each trajectory starts from a fixed default pose and lasts 16--20~s.

The training dataset comprises the following manipulation tasks:
\textit{pick} (move an item from preparation to target plate),
\textit{pour} (tilt a container above the target plate),
\textit{look} (orient the end effector to face the location of any object present, or the plate if no object is present),
\textit{avoid} (orient the end effector sideways, away from any object in the scene).
For each task we collect 60 demonstrations, split 50/50 between Mode A and Mode B (30 per mode).

We also collect two single-mode tasks, \textit{highfive} (approach and touch a human hand when one is present) and \textit{push} (push an item in view to a spot in front of the user's plate), each demonstrated in only one mode---highfive in Mode~A, push in Mode~B---so we can later test whether a motion mode learned on other tasks transfers to a task never demonstrated in that mode (\S\ref{sec:exp_real}). Half of these tasks (pick, pour, push) are precision-demanding direct manipulation, while the other half (look, avoid, highfive) are non-direct manipulation with functional and spatial object/person correlations, together spanning different stages and contact types, from grasping and pushing to contactless orienting and touching. Demonstrations are converted into a Robomimic-format HDF5 dataset; during training we subsample pre-stacked chunks with a stride of 10 to reduce redundancy between consecutive samples.

\subsection{Stage 1 Training: Spatiotemporal Action Tokenizer}

\begin{figure*}[t]
\centering
\includegraphics[width=0.99\textwidth]{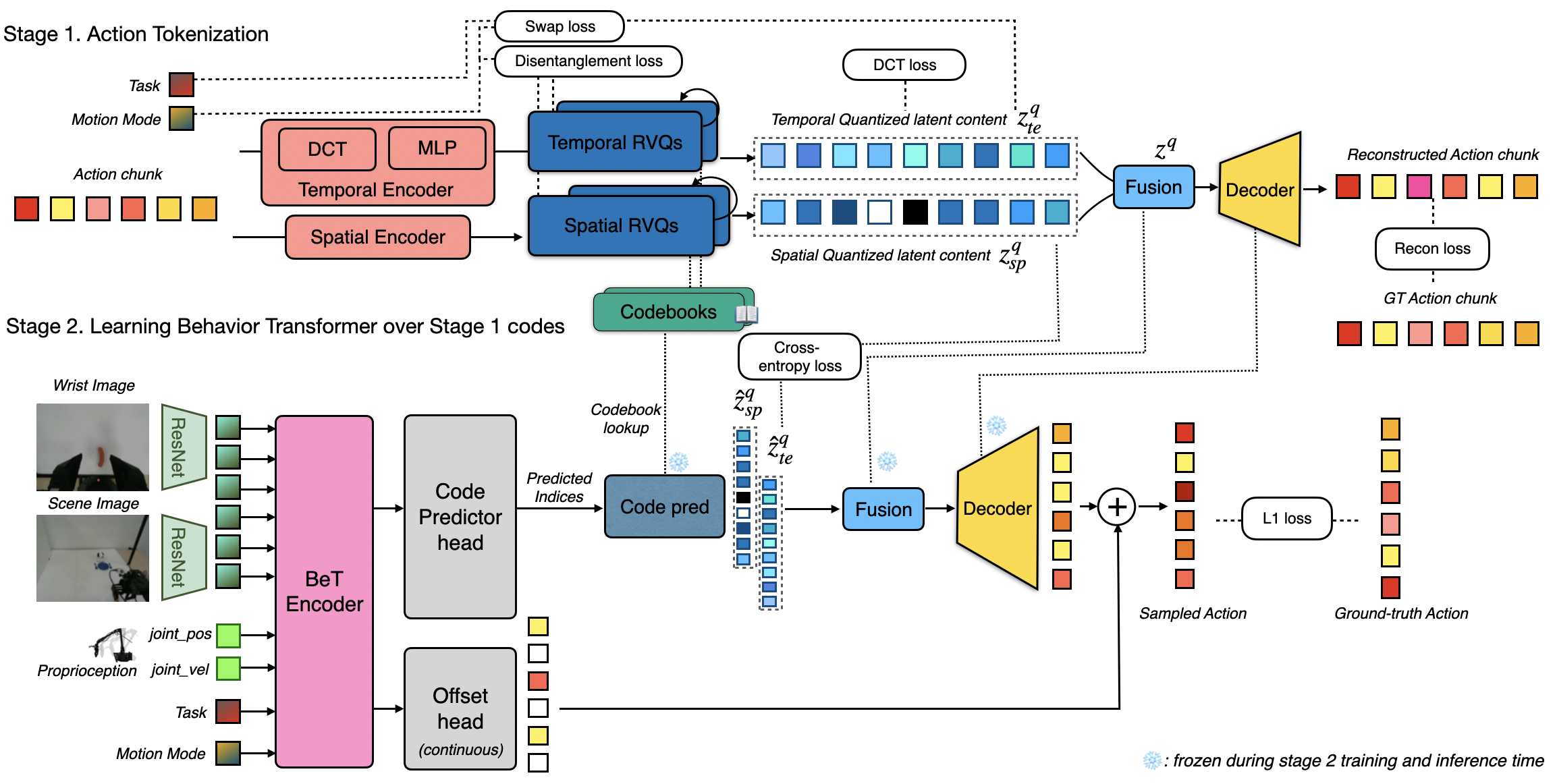}
\caption{MoMo training overview. \textbf{(a) Stage 1:} spatial (1D-conv) and temporal (DCT+MLP) encoders, each quantized by an independent RVQ stack and recombined via a Fusion module into a shared decoder; trained with reconstruction, DCT, disentanglement (adversarial MI + contrastive), and swap losses. \textbf{(b) Stage 2:} a transformer predicts code indices from images, proprioception, task, and motion mode (cross-entropy loss) over the frozen Stage-1 codebooks; a continuous offset head corrects the decoder output (L1 loss).}
\label{fig:arch}
\end{figure*}

Stage~1 is an action tokenizer that converts a chunk $\mathbf{a} \in \mathbb{R}^{100 \times 7}$ into a sequence of discrete codes via two parallel residual VQ stacks: a \emph{spatial} stack and a \emph{temporal} stack. The architecture is shown in Figure~\ref{fig:arch}(top). The two stacks are trained jointly with a shared decoder.

\subsubsection{Architectural overview} The chunk enters two encoders simultaneously. The spatial encoder operates directly on the raw joint-space waveform and produces a temporally compressed latent $\mathbf{z}_\text{sp} \in \mathbb{R}^{K \times D}$ at $K = 20$ positions of dimension $D = 256$. The temporal encoder first computes the DCT of joint deltas, retains the first $16$ low-frequency coefficients per joint, and projects the result through a multilayer perceptron (MLP) to a temporally aligned latent $\mathbf{z}_\text{te} \in \mathbb{R}^{K \times D}$. Each latent stream then feeds an independent RVQ stack. The two quantized streams $\mathbf{z}^q_\text{sp}, \mathbf{z}^q_\text{te}$ are concatenated and passed through a fusion MLP to produce a single $D$-dimensional latent which the decoder upsamples back to a reconstructed chunk $\hat{\mathbf{a}} \in \mathbb{R}^{100 \times 7}$. An auxiliary DCT-prediction head $h_\text{DCT}$ reads the temporal quantized stream $\mathbf{z}^q_\text{te}$ and is supervised to recover the DCT target, ensuring the temporal codes retain frequency-domain information about dynamics. This split encodes a structural prior about manipulation trajectories, detailed alongside each encoder below.

\subsubsection{Spatial encoder}
The spatial encoder $E_\text{sp}$ is a 1D convolutional stack mapping $\mathbf{a} \in \mathbb{R}^{100 \times 7}$ to a latent stream $\mathbf{z}_\text{sp} \in \mathbb{R}^{K \times D}$ with $K = 20$ temporal positions and $D = 256$ channels. The 5$\times$ temporal downsampling lets each latent position summarize a 0.4-second window of joint motion. Because its input is the raw joint-space waveform, i.e., absolute joint positions, the \textbf{spatial encoder naturally captures path geometry, contact, and end-effector targets}.

\subsubsection{Temporal encoder}
The temporal encoder $E_\text{te}$ first computes joint deltas $\Delta q_t = q_{t+1} - q_t$ for $t = 1,\ldots,T-1$, then applies the type-II discrete cosine transform along time and retains the first $C = 16$ frequency coefficients per joint:
\begin{equation}
\hat{\Delta q}^{(j)}_c = \sum_{t=1}^{T-1} \Delta q_t^{(j)} \cos\!\left[\frac{\pi c}{T-1}\left(t-\tfrac{1}{2}\right)\right], \quad c \in \{0,\ldots,C-1\}.
\end{equation}
The resulting feature vector $\hat{\Delta q} \in \mathbb{R}^{C \cdot 7}$ is projected by a 2-layer MLP into $\mathbf{z}_\text{te} \in \mathbb{R}^{K \times D}$. The use of action \emph{deltas} rather than absolute pose emphasizes velocity and acceleration structure, which is the dimension along which our two motion modes differ most. Because its input contains only the time-frequency content of joint velocities and no absolute pose, the \textbf{temporal encoder naturally captures pace, smoothness, and rhythmic structure rather than path geometry}.

\subsubsection{Residual vector quantization}
Each latent stream is quantized by an independent residual VQ stack~\cite{lee2021soundstream,defossez2022encodec}. A single VQ level performs nearest-neighbor codebook lookup with exponential moving average (EMA) codebook updates:
\begin{equation}
i_k^{(\ell)} = \arg\min_n \|r_k^{(\ell)} - e_n^{(\ell)}\|_2, \qquad z_{q,k}^{(\ell)} = e_{i_k^{(\ell)}}^{(\ell)},
\end{equation}
where $r_k^{(\ell)}$ is the residual at level $\ell$ for latent position $k$ and $e_n^{(\ell)} \in \mathbb{R}^D$ are the codebook entries. Residuals propagate through levels via $r_k^{(\ell+1)} = r_k^{(\ell)} - z_{q,k}^{(\ell)}$. The spatial stack uses $L_\text{sp} = 4$ levels with codebook sizes $[64, 512, 512, 512]$; the deliberately small first codebook acts as a content bottleneck. The temporal stack uses $L_\text{te} = 4$ levels of size 256 each. Codebook updates are deferred until after the optimizer step so that contrastive gradients can flow to codebook entries via the residual chain~\cite{zargarbashi2026vqstyle}.

\subsubsection{Decoder fusion}
The quantized streams are concatenated and passed through a fusion MLP:
\begin{equation}
\mathbf{z}_\text{q} = \text{LayerNorm}(\text{GELU}(W_\text{f} [\mathbf{z}_\text{sp}^q \,;\, \mathbf{z}_\text{te}^q])), \quad W_\text{f} \in \mathbb{R}^{D \times 2D},
\end{equation}
followed by a 1D upsampling decoder $\psi$ that produces the reconstructed chunk $\hat{\mathbf{a}} = \psi(\mathbf{z}_\text{q})$.

\subsubsection{Training objective}

Beyond the architectural prior, the training objective actively encourages the intended task/motion-mode split. The Stage~1 loss combines seven terms:
\begin{equation}
\begin{aligned}
\mathcal{L} = \mathcal{L}_\text{recon} &+ \lambda_\text{DCT} \mathcal{L}_\text{DCT} + \lambda_\text{swap-task} \mathcal{L}_\text{swap-task} \\
&+ \lambda_\text{swap-mode} \mathcal{L}_\text{swap-mode} + \lambda_\text{con} \mathcal{L}_\text{con} + \lambda_\text{MI} \mathcal{L}_\text{MI} + \mathcal{L}_\text{commit}.
\end{aligned}
\end{equation}
$\mathcal{L}_\text{recon}$ is L1 reconstruction and $\mathcal{L}_\text{DCT}$ pressures the temporal codes to retain DCT structure. The swap loss ($\mathcal{L}_\text{swap-task}$, $\mathcal{L}_\text{swap-mode}$) decodes a base chunk's spatial codes together with a matched donor's temporal codes (same task, opposite mode) and pushes the result toward the base's path and the donor's frequency content, respectively. The disentanglement loss ($\mathcal{L}_\text{con}$, $\mathcal{L}_\text{MI}$) follows~\cite{zargarbashi2026vqstyle} in combining a residual contrastive term on the temporal codes with an adversarial mutual-information (MI) penalty on the spatial latent to separate task from motion mode in the latent space. $\mathcal{L}_\text{commit}$ is the standard per-quantizer commitment loss summed across both stacks.

\subsection{Stage 2 Training: Learning Behavior Transformer over Stage 1 Codes}

Stage~2 is a behavior-cloning transformer that, conditioned on the current observation, predicts the discrete codes that Stage~1 would assign to the next 2-second action chunk (Figure~\ref{fig:arch}(bottom)). Stage~1 weights---encoders, codebooks, fusion module, and decoder---are frozen during Stage~2 training, so only the transformer and a continuous offset head are learned. By freezing Stage~1, we preserve the spatial/temporal split it learned, treat the resulting code vocabulary as a stable, reusable interface, and reduce Stage~2 to learning a policy over a finite code space rather than a continuous action manifold.

\subsubsection{Architecture} The transformer encoder ingests visual features from the front and wrist cameras (ResNet-34), current proprioception, a task-identifier embedding, and an embedding of the continuous motion-mode scaler $s \in [0,1]$. The decoder has one classification head per codebook level across both stacks, plus a continuous offset head that corrects for residual quantization error. At inference, the predicted indices are looked up in the frozen Stage-1 codebooks, fused, and decoded into an action chunk, with the offset added on top.

\subsubsection{Training objective} Cross-entropy (CE) on the per-codebook indices plus an L1 loss on the offset, with equal weights:
\begin{equation}
\mathcal{L}_\text{Stage2} = \frac{1}{K(L_\text{sp}+L_\text{te})}\sum_{k,\ell} \text{CE}\!\left(p_k^{(\ell)}, i_k^{(\ell)}\right) + \|\mathbf{o} - \mathbf{o}^*\|_1,
\end{equation}
where $p_k^{(\ell)}$ is the predicted code distribution and $i_k^{(\ell)}$ the ground-truth Stage-1 index at position $k$, level $\ell$, and $\mathbf{o}$, $\mathbf{o}^*$ are the predicted and target offsets.

\subsection{Inference}

At test time, the visual and proprioceptive observations, the task identifier, and a user-controlled motion-mode scaler $s$ enter the transformer. The task identifier selects the functional skill, while $s$ modulates its execution quality, making the two conditioning roles separately addressable at inference time. The argmax of each classification head selects the discrete code at every level; the corresponding code embeddings are summed within each stack, fused, and decoded to obtain the action chunk; the offset head adds a continuous correction. To stabilize execution at 50~Hz, we use the temporal-aggregation strategy of Zhao et al.~\cite{zhao2023learning} that averages overlapping predicted chunks.

\section{Experiments}

We evaluate whether motion mode behaves as a reusable behavioral factor in closed-loop real-robot control: whether changing the scaler produces consistent, perceptible changes across tasks, supports interpolation between prototypes, and transfers to held-out task--mode pairs while preserving task success.

\subsection{Evaluation Protocol}
\label{sec:eval_protocol}

We evaluate the trained policy on the same six tasks: four trained on both modes (pick, pour, look, avoid) and two transfer tasks trained on a single mode (push: Mode~B; highfive: Mode~A). Each is rolled out $N=20$ times at three scaler values $s \in \{0.1, 0.5, 0.9\}$ (Mode~B, interpolated, Mode~A), which for the transfer tasks includes the mode never seen in training. Rollouts use varied initial conditions ($\pm 5$~cm object position and orientation) and log the executed joint trajectory at 50~Hz, the front and wrist camera streams, the predicted spatial/temporal codebook indices, and the scaler $s$.

For each rollout segment we compute three motion-dynamics metrics: \emph{joint speed} (mean joint displacement per step, rad/s), \emph{joint acceleration} (mean magnitude of the second difference of joint positions, rad/s$^2$), and \emph{approach pitch} (end-effector gaze direction from FK, in degrees, positive = downward-pointing).
We complement these with a blinded human evaluation (\S\ref{sec:exp_subjective}): three expert raters independently scored every rollout video on \emph{style adhesion} (0--10, Mode~B to Mode~A) and \emph{task completion} (0--10, no motion to clean success), blind to the requested scaler and to whether a cell was a transfer cell.

\subsection{Objective Results}
\label{sec:exp_real}

\begin{figure}[t]
\centering
\includegraphics[width=\columnwidth]{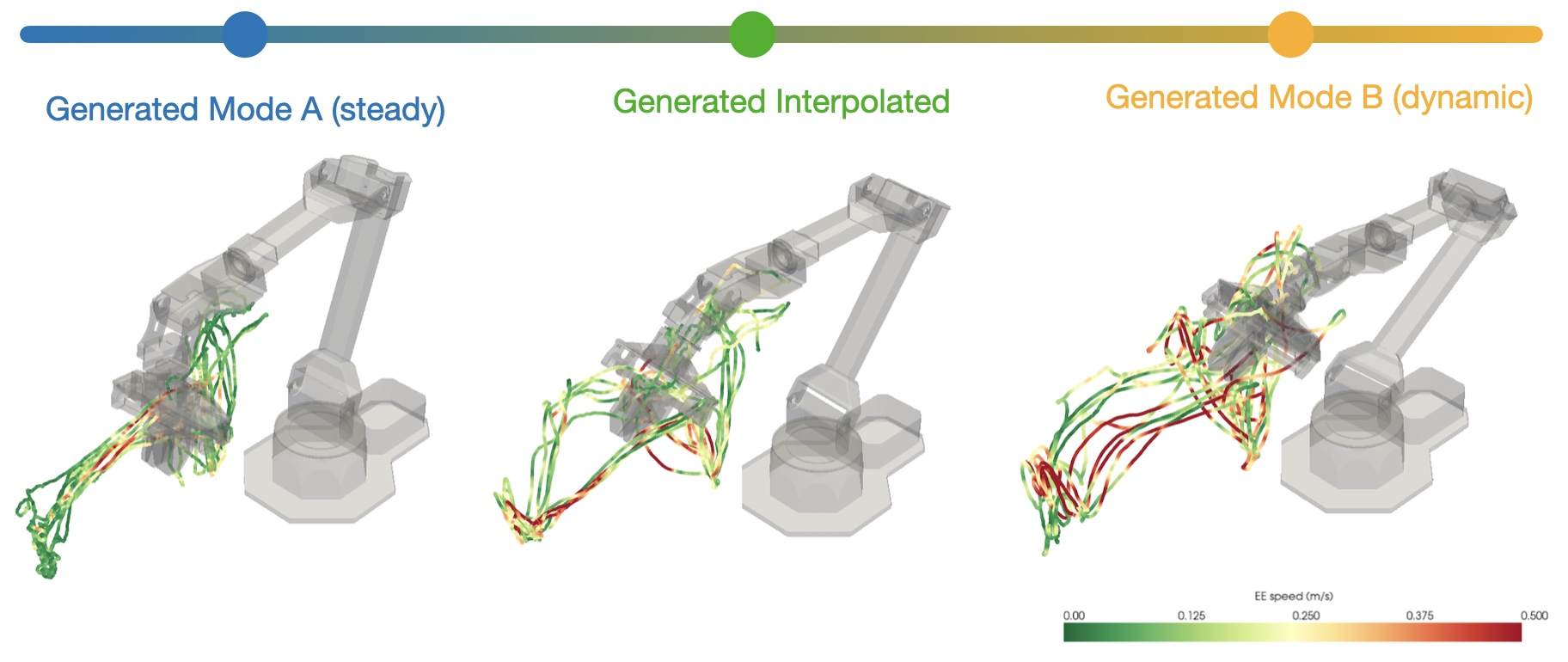}
\caption{End-effector trajectories at each motion-mode scaler (10 rollouts each): Mode A, Interpolated, Mode B. Color encodes instantaneous speed (green = slow, red = fast, clipped at $0.5$ m/s). Mode A paths are tight and uniformly slow; Mode B paths are more spatially varied with pronounced fast segments; Interpolated falls between.}
\label{fig:generated_traj_visual}
\end{figure}

Figure~\ref{fig:generated_traj_visual} qualitatively compares the two endpoint modes and the intermediate condition on the pour task: each panel overlays the end-effector paths of ten representative rollouts at a fixed scaler.

\begin{figure}[t]
\centering
\includegraphics[width=\columnwidth]{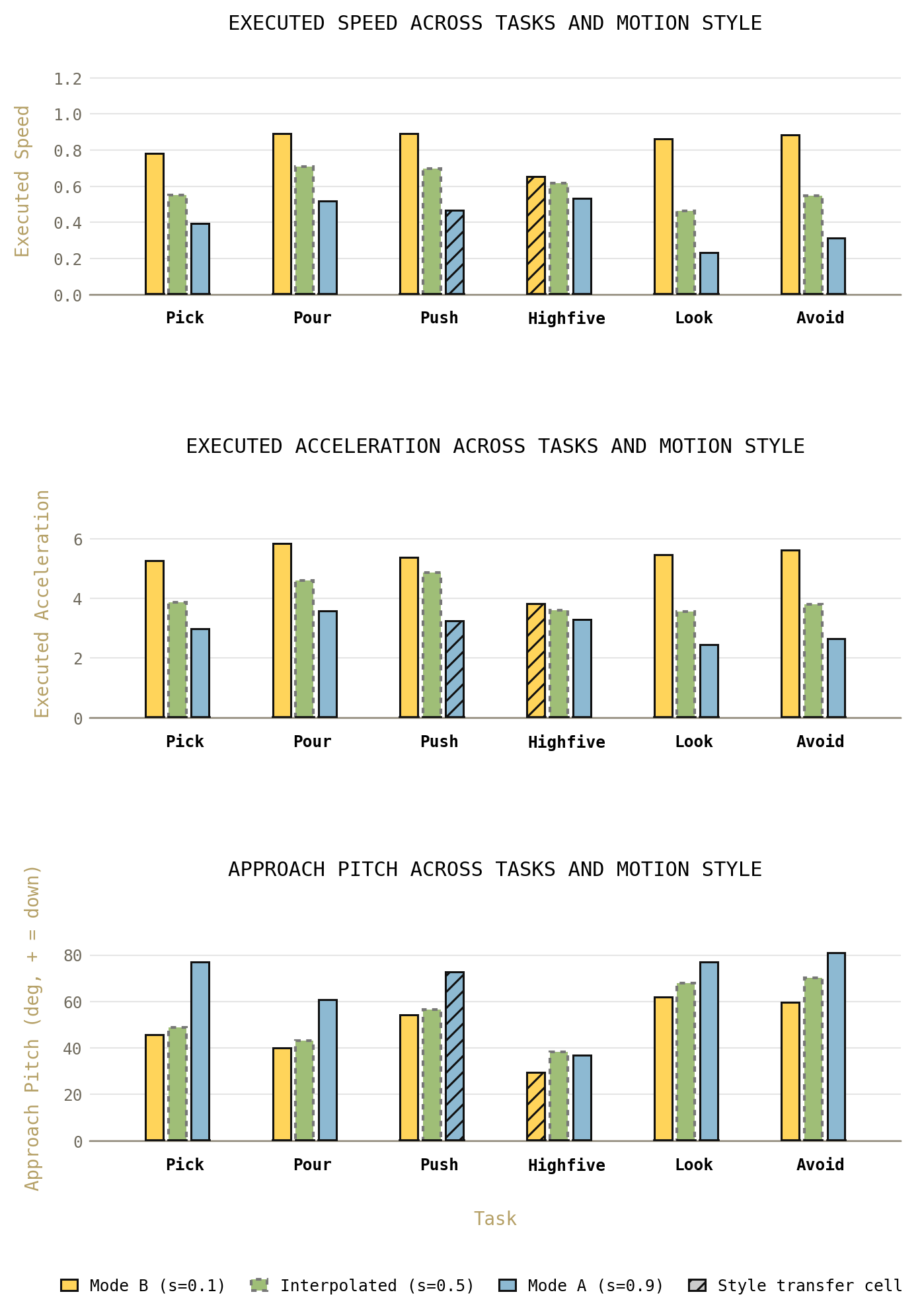}
\caption{Per-segment dynamics by (task, scaler): executed joint speed, executed joint acceleration, and approach pitch. Six tasks $\times$ three scalers: \textbf{Mode B} (yellow), \textbf{Interpolated} (green), \textbf{Mode A} (blue); hatched cells are the two motion-mode transfer conditions (push at Mode A, highfive at Mode B). Every metric separates Mode A from Mode B on trained tasks, with the intermediate scaler falling between.}
\label{fig:real_robot_grouped_metrics}
\end{figure}

\paragraph{In-distribution mode conditioning} Figure~\ref{fig:real_robot_grouped_metrics} reports per-cell mean values of the three motion-dynamics metrics per (task, scaler) cell, computed over all marked rollout segments. On the four trained tasks, all three metrics separate Mode A from Mode B cleanly: Mode A keeps the gripper $15$--$25^\circ$ further down on average and is steadier and slower than Mode B, with the intermediate $s=0.5$ bar reliably falling between them on every metric.

\paragraph{Intermediate mode condition} We test whether the three evaluated scaler values produce ordered changes by computing, for each task, the Spearman rank correlation between the requested $s$ and the commanded-action metrics \texttt{speed\_mean} and \texttt{accel\_mean} (Table~\ref{tab:interp_spearman}); the convention is that high $s$ corresponds to Mode A (slower, steadier), so a negative correlation is the desired outcome.

\begin{table}[h]
\centering
\small
\caption{Spearman $\rho$ between scaler $s$ and commanded-action dynamics ($N=20$ segments per task).}
\label{tab:interp_spearman}
\begin{tabular}{lrr}
\toprule
Task & $\rho$(\texttt{speed\_mean}) & $\rho$(\texttt{accel\_mean}) \\
\midrule
look  & $-0.933$ & $-0.943$ \\
avoid & $-0.900$ & $-0.932$ \\
pour  & $-0.914$ & $-0.926$ \\
pick  & $-0.879$ & $-0.933$ \\
\midrule
push (transfer)      & $-0.777$ & $-0.736$ \\
highfive (transfer)  & $-0.647$ & $-0.867$ \\
\bottomrule
\end{tabular}
\end{table}

All four trained tasks show strong, highly significant negative correlations ($p < 10^{-6}$ for all eight tests), \textbf{indicating that execution dynamics change monotonically across the three evaluated scaler settings, with the intermediate condition falling between the endpoint modes.}

\paragraph{Mode transfer} The push and highfive cells test the central reuse hypothesis: whether a motion mode learned from other tasks can modulate a task observed in only \emph{one} mode. Both shift in the requested direction, but with different magnitudes. The hatched cells in Figure~\ref{fig:real_robot_grouped_metrics} mark the two transfer conditions (push at $s=0.9$, highfive at $s=0.1$). Push, trained only in Mode~B, moves speed and acceleration toward Mode~A, with speed entering the trained-task Mode~A range and approach pitch tilting further downward. Highfive, trained only in Mode~A, shifts toward Mode~B but remains between the endpoint ranges on speed and acceleration. Both also retain their demonstrated-mode profiles at the trained endpoint. Table~\ref{tab:interp_spearman} confirms statistically significant monotonic correlations between scaler and dynamics ($p < 5\times10^{-3}$ for all four tests), though weaker than on the trained tasks. \textbf{Together, these results show that MoMo can generate trajectories for unseen task--mode pairs whose quantitative motion metrics shift toward values associated with the requested mode on other tasks.}

\subsection{Latent Space Visualization and Insights}
\label{sec:latent_viz}

\begin{figure}[t]
\centering
\includegraphics[width=0.5\textwidth]{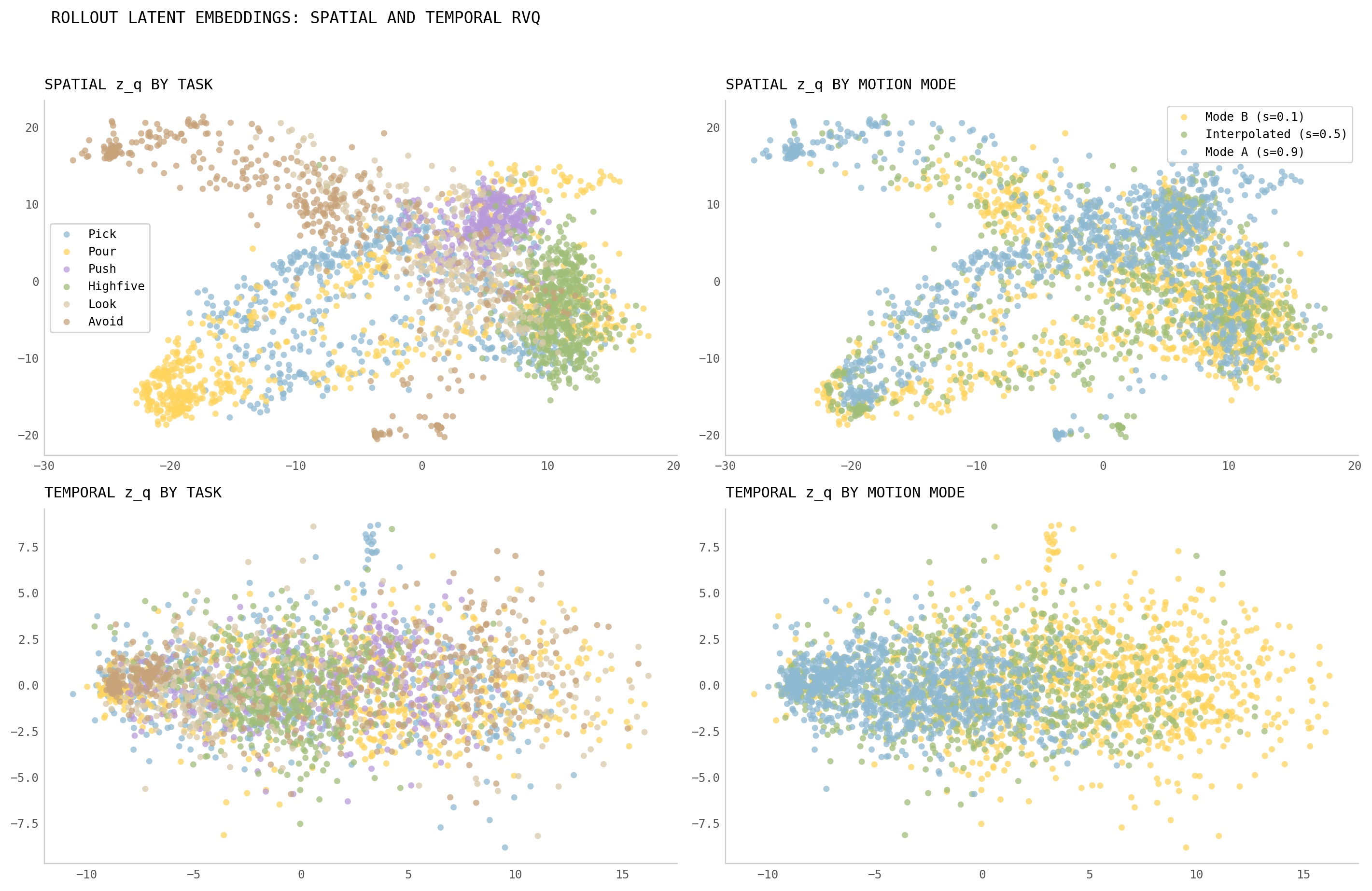}
\caption{Quantized latents $\mathbf{z}^q_\text{sp}$ (top) and $\mathbf{z}^q_\text{te}$ (bottom), projected via PCA; colored by task (left) and scaler $s$ (right). Spatial codes cluster by task and are interleaved across modes; temporal codes form a clean Mode B / Mode A band and overlap across tasks. The Stage~1 decomposition is preserved at deployment.}
\label{fig:latent_pca}
\end{figure}

We project the recovered Stage~1 quantized latents $\mathbf{z}^q_\text{sp}, \mathbf{z}^q_\text{te}$ from the same rollouts into 2D with PCA, colored by task and by scaler $s$ (Figure~\ref{fig:latent_pca}). Spatial codes separate by task and are largely insensitive to motion mode: per-task clusters form along the principal axes, with different scalers interleaved within each task. Temporal codes show the complementary structure, forming a clean Mode~B--Mode~A band along PC1 while all six tasks overlap. The Stage~1 decomposition therefore persists through Stage~2 at deployment. Because mode-structured temporal codes overlap across tasks while spatial codes retain task clusters, this organization is consistent with behavioral variation being reused across task-specific action structure.

\begin{figure}[t]
\centering
\includegraphics[width=0.5\textwidth]{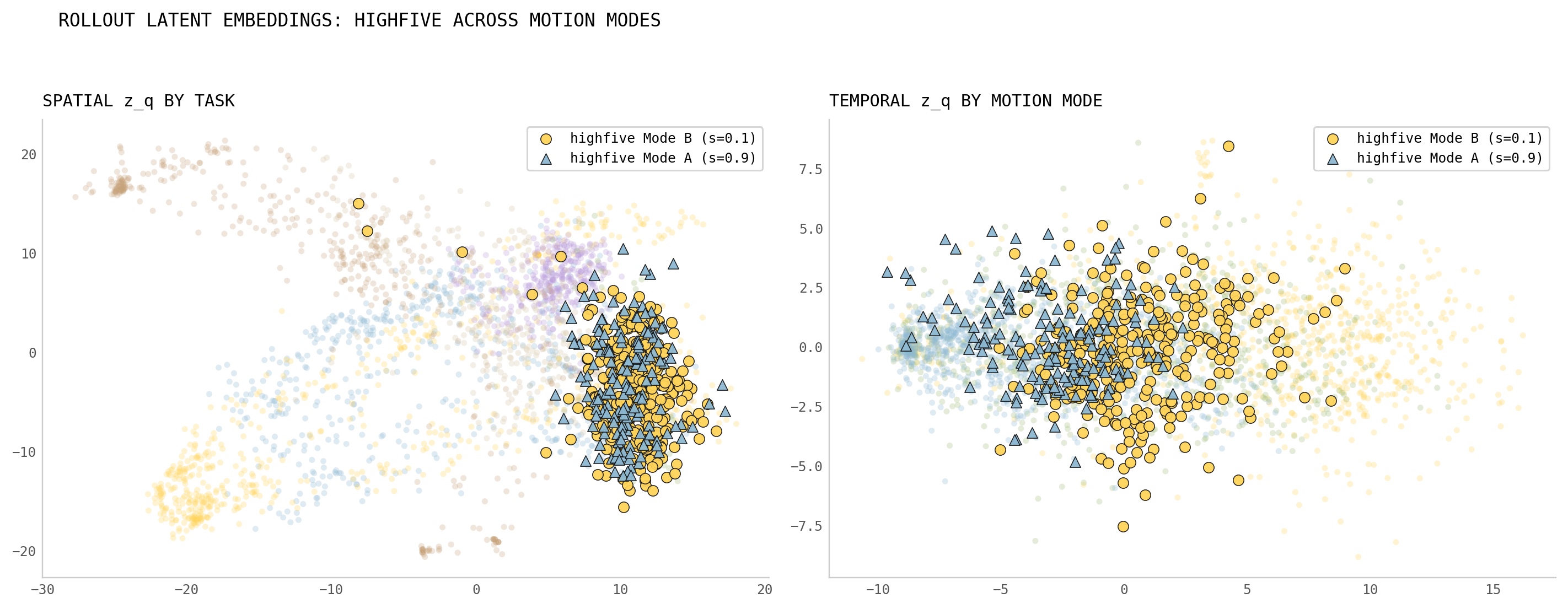}
\caption{Highfive rollout latents at Mode B ($s=0.1$, yellow circles, transferred) and Mode A ($s=0.9$, blue triangles, trained), projected onto the training-latent PCA (background, colored as in Figure~\ref{fig:latent_pca}). \textbf{Left:} spatial $\mathbf{z}^q_\text{sp}$---both modes stay confined to the same highfive task cluster, regardless of requested mode. \textbf{Right:} temporal $\mathbf{z}^q_\text{te}$---Mode~B is pushed toward the Mode~B region of the training distribution even though highfive was never demonstrated there.}
\label{fig:highfive_modes_latent}
\end{figure}

The same decomposition holds for the two single-mode tasks. Although push was observed only in Mode~B and highfive only in Mode~A, their spatial codes form separate task-specific clusters, while their trained-mode temporal codes occupy the corresponding mode regions. For the unseen (highfive, Mode~B) pair, the spatial code remains confined to the highfive cluster regardless of requested mode, while the temporal code shifts toward the Mode~B training region even though highfive was never demonstrated there (Figure~\ref{fig:highfive_modes_latent}). \textbf{Transfer is therefore expressed primarily by shifting the temporal latent while the spatial latent remains task-constrained, a compositional pattern consistent with reusing behavioral variation across task-specific structure.} It also explains the weaker speed response: the requested temporal code changes the dynamics, but highfive's short-range spatial geometry bounds the speed that the fused decoder can express.

\subsection{Human Subject Evaluation}
\label{sec:exp_subjective}

\begin{figure*}[t]
\centering
\includegraphics[width=\textwidth]{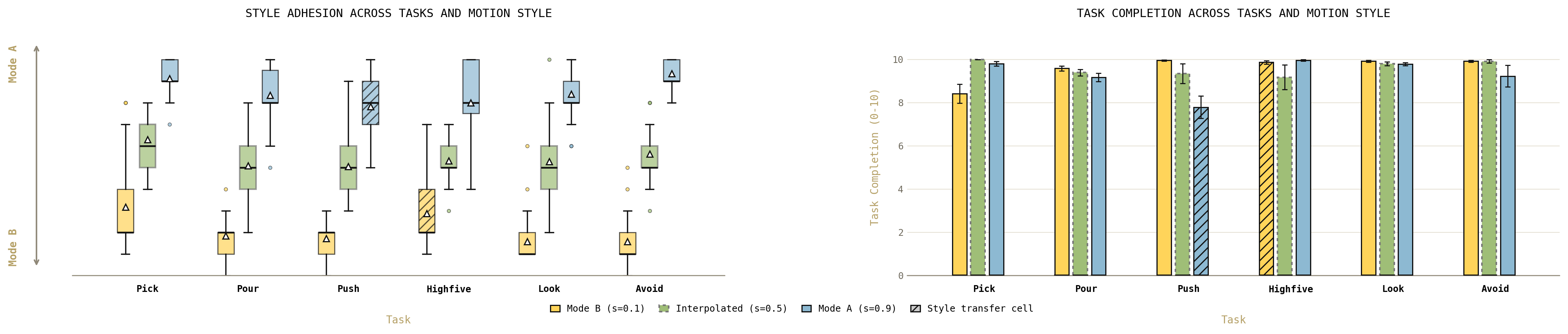}
\caption{Subjective Likert evaluation across six tasks $\times$ three scalers. \textbf{Left:} mean style adhesion, monotonic in the requested scaler on every task, including the two motion-mode transfer cells (hatched). \textbf{Right:} mean task completion (0 = complete failure, 10 = fully successful). Color: Mode~B yellow, Interpolated green, Mode~A blue. Error bars: SEM across raters/rollouts.}
\label{fig:human_eval_summary}
\end{figure*}

On every trained task, style-adhesion ratings are monotonic in the requested scaler---Mode~B rated as dynamic, Mode~A as steady, and the interpolated condition falling consistently between them (Figure~\ref{fig:human_eval_summary}, left). \textbf{The same ordering holds on the two motion-mode transfer cells, confirming that motion mode is perceived consistently by human raters} even on tasks demonstrated in only one mode: highfive is rated correctly despite its objective speed/acceleration metrics barely separating across $s$, suggesting raters pick up on dynamics differences (smoothness, hesitation, gripper choreography) that integrated joint speed alone does not capture. This stable perceptual ordering across tasks suggests that the condition retains a shared behavioral meaning rather than becoming task-specific.

Task completion remains near ceiling across nearly all cells (Figure~\ref{fig:human_eval_summary}, right), with one exception: push at the held-out Mode~A, where the trajectory is correctly slowed but the contact phase becomes less reliable. \textbf{Task performance is otherwise preserved even on the mode-transfer tasks.}

\section{Discussion}
\label{sec:disc}

This paper presents \textbf{MoMo}, a two-stage imitation-learning framework that tokenizes robot action chunks into spatial and temporal residual-VQ streams and conditions the resulting policy on motion mode. We view motion mode as a reusable behavioral modulation factor rather than task-specific trajectory variation: it changes how a task is executed without specifying the task itself. Across six real-robot tasks, MoMo interpolates between two prototype dynamics and transfers a requested mode to tasks that never demonstrated it, shifting execution toward the request while retaining task-specific spatial structure and largely preserving task success. The objective metrics, latent projections, and blinded human ratings therefore provide evidence of a limited but meaningful form of compositional task--mode generalization.

MoMo encourages this factorization architecturally---absolute joint trajectories in the spatial branch, DCT features of joint deltas in the temporal branch---and through the contrastive, adversarial, and swap losses that push the two streams to carry complementary information. The result should not be read as motion mode living inside the temporal codebook, however: the final action representation fuses both streams, and motion mode has geometric correlates that the spatial branch can absorb. A more cautious interpretation is that MoMo learns a fused action-token space in which the temporal stream serves as a reusable behavioral modulation channel.

The factorization remains partial. Motion mode carries geometric correlates that the spatial encoder partly absorbs, limiting the magnitude of transfer: highfive's short-range spatial code bounds the speed expressed under Mode~B, and push at the held-out Mode~A shifts in the correct direction but becomes less reliable at contact. Raters nonetheless identified highfive's transferred mode correctly even though its speed and acceleration barely separated across $s$, suggesting that our scalar dynamics metrics under-measure the behavioral change the policy produces. Offline analysis points the same way: the temporal codes transfer summary statistics of dynamics---mean speed, mean acceleration, smoothness---more faithfully than the full DCT spectrum. The evaluation is also narrow: one embodiment, six tasks, $N=20$ rollouts per cell, initial object positions fixed within $\pm 5$~cm, two prototype modes on a one-dimensional scaler, and three expert raters rather than naive participants. These results therefore constitute initial evidence for task--mode factorization, not a fully disentangled representation, and they motivate richer behavioral factors, stronger swap supervision, and larger task distributions that expose more combinations of task, object, environment, and motion mode.

Taken together, the cross-task latent organization and OOD transfer suggest a way to think about behavioral latents in manipulation policies. A behavioral latent need not be an executable primitive. It can instead be a compact modulation factor that composes with task-specific action structure. Under this view, motion mode is one measurable instance of a broader class of reusable behavioral factors. This perspective may become especially useful as robot policies are pretrained over larger skill distributions and then adapted with limited task-specific data: lightweight adaptation could tune not only what task a robot performs, but also which behavioral factors should modulate that task.

This framing also has implications for interactive deployment. Because motion mode is a run-time conditioning input, an operator or higher-level policy can select $s$ per episode as context changes, without retraining or changing the task identifier. Future work may replace this manual selection with an LLMs-based high-level policy that reads interaction and task context and orchestrates motion-mode shifts on its own, allowing robots to adapt their behavior during interaction rather than only before task execution.

\bibliographystyle{IEEEtran}
\bibliography{reference}

\end{document}